\documentclass{article}

\PassOptionsToPackage{numbers, compress}{natbib}

\usepackage[final]{workshop_neurips_2023}

\usepackage[utf8]{inputenc} 
\usepackage[T1]{fontenc}    
\usepackage{hyperref}       
\usepackage{url}            
\usepackage{booktabs}       
\usepackage{amsfonts}       
\usepackage{nicefrac}       
\usepackage{microtype}      
\usepackage{xcolor}         

\usepackage{graphicx}
\usepackage{amsmath}
\usepackage{amssymb}
\usepackage{booktabs}
\usepackage{multirow}
\usepackage{tikz}
\usepackage{pgfplots}
\usepackage{subfig}
\usepackage{tabularx}
\usepackage{bm}

\title{Inferring Latent Class Statistics from Text\\for Robust Visual Few-Shot Learning}

\author{Yassir Bendou \\ IMT Atlantique, Brest, France \\ \texttt{yassir.bendou@imt-atlantique.fr} \And Vincent Gripon \\ IMT Atlantique, Brest, France \\ \texttt{vincent.gripon@imt-atlantique.fr} \And Bastien Pasdeloup \\ IMT Atlantique, Brest, France \\ \texttt{bastien.pasdeloup@imt-atlantique.fr} \And Giulia Lioi \\ IMT Atlantique, Brest, France \\ \texttt{giulia.lioi@imt-atlantique.fr} \And Lukas Mauch \\ Sony Europe, B.V. \\ Stuttgart Laboratory 1, Germany \\ \texttt{lukas.mauch@sony.com} \And Fabien Cardinaux \\ Sony Europe, B.V. \\ Stuttgart Laboratory 1, Germany \\ \texttt{fabien.cardinaux@sony.com} \And Ghouthi Boukli Hacene \\ Sony Europe, B.V. \\ Stuttgart Laboratory 1, Germany \\ Mila, Montreal, Canada\\ \texttt{ghouthi.bouklihacene@sony.com}
}

\begin{document}

\maketitle

\author{First Author\\
Institution1\\
Institution1 address\\
{\tt\small firstauthor@i1.org}
\and
Second Author\\
Institution2\\
First line of institution2 address\\
{\tt\small secondauthor@i2.org}
}
\maketitle

\begin{abstract}
   In the realm of few-shot learning, foundation models like CLIP have proven effective but exhibit limitations in cross-domain robustness especially in few-shot settings. Recent works add text as an extra modality to enhance the performance of these models. Most of these approaches treat text as an auxiliary modality without fully exploring its potential to elucidate the underlying class visual features distribution. In this paper, we present a novel approach that leverages text-derived statistics to predict the mean and covariance of the visual feature distribution for each class. This predictive framework enriches the latent space, yielding more robust and generalizable few-shot learning models. We demonstrate the efficacy of incorporating both mean and covariance statistics in improving few-shot classification performance across various datasets. Our method shows that we can use text to predict the mean and covariance of the distribution offering promising improvements in few-shot learning scenarios.
\end{abstract}

\section{Introduction}
\label{sec:intro}

Few-shot learning has experienced a significant transformation due to the advent of foundation models. These large pre-trained vision models, such as CLIP~\cite{clip} and DINO~\cite{dino} have shown remarkable performance in various applications but still face challenges in terms of robustness and performance consistency in cross-domain settings. Our work aims to address these limitations by exploring the synergistic use of text as an additional modality.

Text serves as a rich source of semantic information that can offer insights into the class definition, clarify concepts, and even outline the boundaries of what a class entails. While some previous works have added text as a supplemental feature for classification, we aim at proposing a method exploring more nuanced ways of aligning text and images to fully exploit semantic knowledge.

Our primary focus is to examine whether statistics such as the covariance of the distribution of visual features for a class can be inferred from text. This is especially pertinent in few-shot learning scenarios, where we often lack comprehensive data to understand the class distribution fully. By being able to predict statistics about the feature distributions from text, we can potentially improve the robustness and generalizability of few-shot classifiers. Specifically, we explore:
\begin{itemize}
    \item If we can predict the mean and covariance of visual feature distribution from text.
    \item The utility of such predictions in enhancing the performance of few-shot learning models.
\end{itemize}

By focusing on these questions, our work explores the next frontier in few-shot learning, pushing the boundaries of what is achievable with current foundation models by augmenting them with rich, text-derived statistical information. The code for our experiments is available at: \url{https://github.com/ybendou/fs-text2stats}.
\section{Related Work}
\begin{figure}[!t]
    \centering
    \includegraphics[scale=0.2]{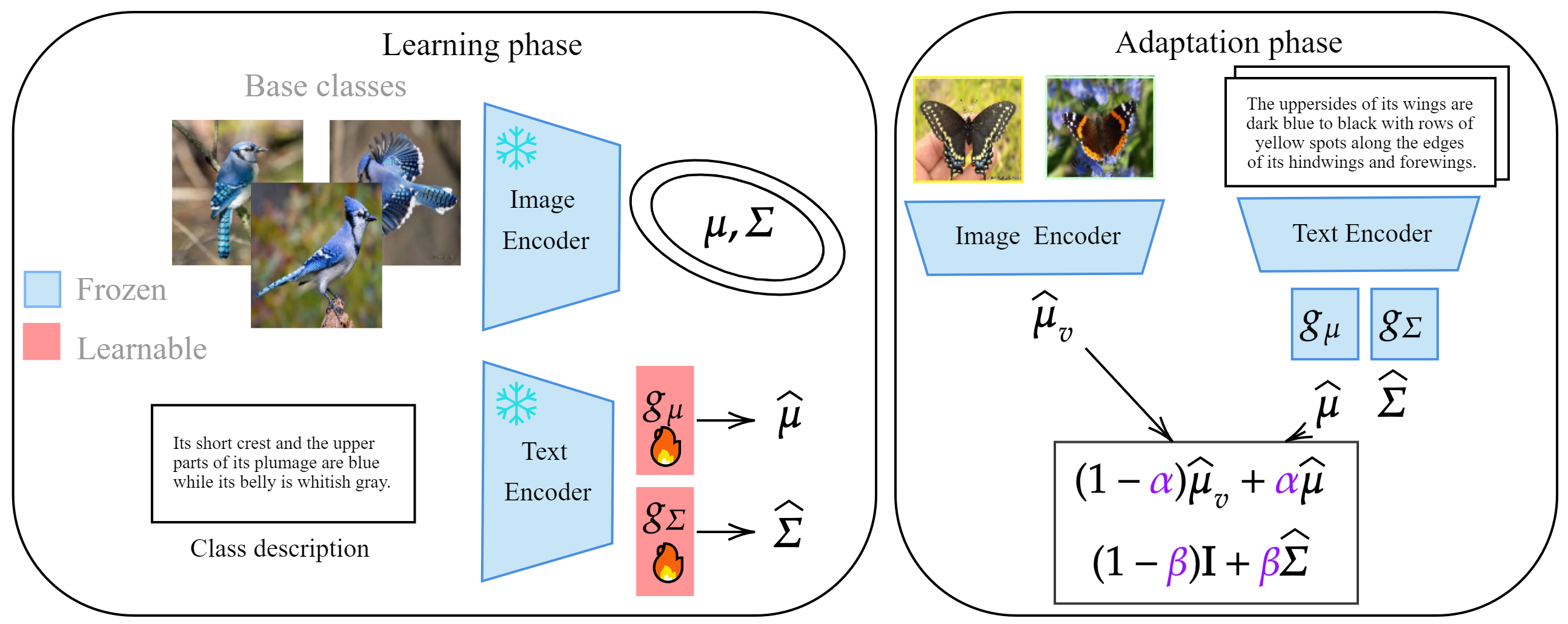}
    \caption{Our pipeline is composed of two phases. We first train two networks to infer the mean and the covariance of the latent visual distribution. Secondly, given a few-shot downstream task, we interpolate the predictions of the mean using the available shots and apply shrinkage for the covariance matrix. We then use these statistics to perform few-shot classification.}
    \label{fig:enter-label}
\end{figure}

\paragraph{Multimodal foundation models}
In recent years, foundation models have increasingly adopted a multimodal supervision paradigm, integrating multiple types of data such as text and images for more effective learning. In the domain of vision, initial research efforts focused on converting web-based image captions into structured formats suitable for supervised learning. These formats included multi-label classification targets and visual n-grams. More contemporary methods, exemplified by CLIP and ALIGN~\cite{align}, employ a contrastive learning framework to map both images and their associated captions into a shared representational space. These methods have shown remarkable capabilities in "zero-shot" learning, demonstrating high performance on downstream tasks without requiring additional fine-tuning.
\paragraph{Text for visual few-shot classification}
Several approaches have benefited from the extensive pre-training provided by models like CLIP, which utilizes a large-scale dataset of image-caption pairs to train robust encoders for both text and images~\cite{clip,klite,boosting}. 
There are many streams of work to adapt text to visual tasks. The first one is prompt learning which aims to learn effective prompts in the presence of limited labeled data~\cite{coop}. These approaches differ in how they optimize for prompt generalization across tasks and classes. For instance, CoCoOp introduces query-based token conditioning for better generalizability~\cite{cocoop}, while PLOT employs an optimal transport framework to combine multiple local prompts~\cite{plot}. Another strand of work, including WiSE-FT and Clip-Adapter, focuses on optimizing the fine-tuning process by introducing lightweight, additive modules to the pre-trained models. These methods vary in the extent to which they leverage the limited shots available for fine-tuning~\cite{wise_ft,clip_adapter, tip_adapter}.

Recent work has examined how text can function as additional data points for training~\cite{cross_modal}. Methods such as DALLE-based shot generation go beyond mere text prompts, employing generated images as additional training samples for classification~\cite{promptCacheGenerate}. Another work trains a VAE conditioned on semantic embeddings to generate additional points in the feature space~\cite{vaetext}. Unlike these approaches, our work considers text as a statistical summary of the class, capturing both mean and covariance.

While previous work has mostly relied on simplistic prompts based solely on class labels, some recent studies have demonstrated the utility of more descriptive class prompts. Our approach extends this line of work by experimenting with both class labels and more informative visual class descriptions~\cite{visual_descriptions}.


\section{Approach}
\label{sec:approach}

In this section, we present our method to predict from text the mean and covariance of a class distribution in the feature space. We split our method in two phases, a learning phase where we learn a mapping from text to these order of moments, and an adaptation phase where we adjust the predictions to the downstream dataset. 
\subsection{Learning Phase}
In the learning phase, our objective is to predict both the mean and covariance of visual features using textual descriptions. Having access to statistics about the visual distribution are important for few-shot learning where only few samples are available to infer the latent distribution. Let $\mathbf{x}$ be an image and $\mathbf{t}$ the text of its associated label $y$. We aim to utilize a pre-trained visual backbone, denoted by $f_{v}$, to extract image features $\mathbf{z}=f_{v}\left(\mathbf{x}\right)$ and a pre-trained text encoder, denoted by $f_{t}$, to extract text features $\mathbf{s}=f_{t}\left(\mathbf{t}\right)$. For text, we either utilize different text contexts such as "a photo of a \{class\}" or we use GPT3~\cite{gpt3} to generate a visual description for each class, which has been used previously in the literature~\cite{promptCacheGenerate} (see Appendix for more details).

Given some semantic features $\mathbf{s}_i$ of a class $c_i$, our target is to accurately estimate the mean and covariance of the visual features $\bm{\mu}_{i}=\mathbb{E}_{\substack{(\mathbf{x},y)\\y=c_i}}\left(f_{v}\left(\mathbf{x}\right)\right)$ and $\mathbf{\Sigma}_{i}=\mathbb{E}_{\substack{(\mathbf{x},y)\\y=c_i}}\left(\left(f_{v}\left(\mathbf{x}\right) - \mu_i\right)\left(f_{v}\left(\mathbf{x}\right) - \mu_i\right)^T\right)$. Given the high-dimensionality of these features, we choose to infer a diagonal covariance matrix.

We employ two mapping networks, $g_{\mu}\left(\mathbf{s},\theta_{\mu}\right)$ and $g_{\Sigma}\left(\mathbf{s},\theta_{\Sigma}\right)$ for predicting the mean and covariance. Leaning on the strength of CLIP for robust textual embeddings, these networks are compactly designed, primarily using a Multi-Layer Perceptrons with 2 layers. The training for these networks is executed on a base dataset (more details in Section~\ref{sec:experiments}). Using the base dataset composed of $N$ classes, we compute the empirical mean and covariance for each class $\Tilde{\bm{\mu}}_i$ and $\Tilde{\mathbf{\Sigma}}_i$ respectively and train the mapping networks using an $L$2 loss on these statistics combined with a regularization term $R$ (\emph{e.g.}, weight decay): $L(\theta_{\bm{\mu}})~=\frac{1}{N}\sum_{i=1}^{N} \|g_{\mu}(\mathbf{s}_i,\theta_{\mu})-\Tilde{\bm{\mu}}_i\|_2 + R\left(\bm{\theta}_{\bm{\mu}}\right)$ and $L(\theta_{\bm{\Sigma}})~=\frac{1}{N}\sum_{i=1}^{N} \|g_{\Sigma}(\mathbf{s}_i,\theta_{\Sigma})-\Tilde{\bm{\Sigma}}_i\|_2 + R\left(\bm{\theta}_{\bm{\Sigma}}\right)$.

\subsection{Adaptation Phase}
Having inferred estimators for the mean and the covariance, integrating them into the classification process would be straightforward, as delineated in Section~\ref{sec:experiments}. Nevertheless, to account for potential domain shifts between training and test datasets, we introduce regularization techniques. Specifically for the mean, similar to~\cite{saga}, we interpolate our estimate using the empirical mean from the few available shots in the downstream task denoted $\hat{\bm{\mu}}_{i,v}$ as follows: $\hat{\bm{\mu}}_i = \left(1-\alpha\right)\hat{\bm{\mu}}_{i,v} + \alpha g_{\mu}\left(\bm{s}_i\right)$. Due to the unreliability of covariance estimates from few-shots, we employ shrinkage, adjusted by a coefficient $\beta$: $\hat{\bm{\Sigma}}_i = \left(1-\beta\right)\mathbf{I} + \beta g_{\Sigma}\left(\bm{s}_i\right)$. In our experiments, we assume that the covariance is diagonal, which means that a simple interpolation in the euclidean space is sufficient. 

\section{Experiments}
\label{sec:experiments}

\subsection{Experimental protocol}
In our experiments, we use two base datasets: ImageNet~\cite{imagenet} and iNaturalist~\cite{inaturalist}. We chose iNaturalist given that it's a hierchical dataset with fine-grained classes (different species that are very similar to each other sometimes), in such a dataset, the covariance matrix plays a crucial role. For the test dataset, we experiment with either a disjoint subset of the base classes or 9 cross-domain datasets.: Caltech~\cite{caltech101}, EuroSAT~\cite{eurosat}, Food~\cite{food101}, Flowers~\cite{oxford_flowers}, SUN397~\cite{sun397}, DTD~\cite{dtd}, Pets~\cite{oxford_pets}, Cars~\cite{stanford_cars} and UCF101~\cite{ucf101}. We extract visual and text features using the pre-trained CLIP ResNet50 which was trained on LAION400M~\cite{LAION400M}, a disjoint dataset from the previously mentionned ones.


\subsection{Mapping networks training}
\begin{center}
\begin{table}[!htbp]
\centering
 \scalebox{0.8}{
  \begin{minipage}{1\linewidth}
\begin{center}
\begin{tabular}{cllllll}
\cmidrule(r){4-7}
& & & \multicolumn{2}{c}{$\bm{\mu}$}        & \multicolumn{2}{c}{$\mathbf{\Sigma}$}     \\
\cmidrule(r){4-5} \cmidrule(r){6-7}
 Base dataset & & Model & val & test & val & test \\

\midrule
\multirow{3}{*}{\rotatebox{0}{iNaturalist}} & \multirow{3}{*}{\rotatebox{90}{ID}} & Baseline  & $1.00\pm0.02$ & $0.97\pm0.02$  & $1.00\pm0.03$             & $0.99\pm0.03$              \\

& & Class labels   & $0.60\pm0.02$  & $0.60\pm0.02$ & $0.66\pm0.03$ & $0.72\pm0.03$  \\

& & Class description & $\mathbf{0.42\pm0.01}$ & $\mathbf{0.42\pm0.01}$ & $\mathbf{0.54\pm0.02}$ & $\mathbf{0.56\pm0.02}$ \\
\cmidrule(r){1-7}
\multirow{6}{*}{\rotatebox{0}{ImageNet}} & \multirow{3}{*}{\rotatebox{90}{ID}} & Baseline           & $1.00\pm0.02$   
 & $0.98\pm0.02$   & $1.01\pm0.05$             & $1.00\pm0.04$ \\
& & Class labels     & $\mathbf{0.44\pm0.02}$  & $\mathbf{0.45\pm0.01}$ & $\mathbf{0.70\pm0.03}$ & $\mathbf{0.74\pm0.04}$  \\
& & Class description  & $0.46\pm0.02$ & $0.47\pm0.01$ & $\mathbf{0.70\pm0.03}$ & $\mathbf{0.74\pm0.04}$\\
\cmidrule(r){2-7}
 & \multirow{3}{*}{\rotatebox{90}{CD}} & Baseline          & $0.99\pm0.03$ &  $1.67\pm0.10$  & $1.04\pm0.07$ & $4.99\pm0.90$ \\
& & Class labels  & $\mathbf{0.43\pm0.03}$ & $\mathbf{1.26\pm0.11}$ & $0.72\pm0.06$   & $\mathbf{3.99\pm0.60}$ \\
& & Class description & $0.45\pm0.02$ & $1.30\pm0.11$ & $\mathbf{0.70\pm0.05}$ & $4.67\pm0.84$\\
\bottomrule
\end{tabular}
\end{center}
\end{minipage}
}
\vspace{1.5mm}
\caption{Mean Squared Error of predictions from ImageNet-1k and iNaturalist. The baseline is the average value over the training set. We experiment with both in-domain (ID) and cross-domain (CD) settings. The results are averaged across 10 different seeds.}
\label{table:MSE_indomain_crossdomain}
\end{table}
\end{center}
\vspace{-1mm}
Let us first show that it is possible to accurately estimate both means and covariances of visual features of class from its label. To this end, we train a model to predict the mean (resp. the covariance) of visual features from the text features of its class label. These models were trained to minimize the MSE on a train set, then tested on a validation set, either in a In-Domain (ID) setting or Cross-Domain (CD) setting. Means and covariances were standardized for this experiment, such that the MSE of the baseline in the ID case is 1 (more details in the Appendix).

We can derive multiple results from this first experiment in Table~\ref{table:MSE_indomain_crossdomain}. First, we observe that in the ID case, our models obtain MSE significantly lower than 1, showing there is indeed information about both the mean and covariance hidden in the text model. Second, we observe that for iNaturalist, using the description of a class gives significantly better results than the class labels which is expected given that iNaturalist often contains uncommon class names. This result does not hold for ImageNet which contains common names. Therefore, for all our experiments, we use class descriptions for iNaturalist and class labels when using ImageNet as a base dataset. 

\subsection{Classification}
\vspace{-0.8cm}
\begin{figure}[htbp!]
\centering
\subfloat[One-class]{\scalebox{0.76}{{
\begin{tikzpicture}
\begin{scope}[xscale=1]

\definecolor{green}{RGB}{0,128,0}
\definecolor{lightgray203}{RGB}{203,203,203}
\definecolor{lightgray204}{RGB}{204,204,204}
\definecolor{orange}{RGB}{255,165,0}
\definecolor{whitesmoke240}{RGB}{240,240,240}

\begin{axis}[
axis background/.style={fill=white},
axis line style={whitesmoke240},
legend cell align={left},
legend style={
  fill opacity=0.8,
  draw opacity=1,
  text opacity=1,
  at={(0.97,0.03)},
  anchor=south east,
  draw=lightgray204,
  fill=white
},
tick align=outside,
tick pos=left,
unbounded coords=jump,
x grid style={lightgray203},
xlabel={number of shots},
xmajorgrids,
xmin=-1, xmax=21,
xtick style={color=black},
y grid style={lightgray203},
ylabel={AUROC $\uparrow$ (\%)},
ymajorgrids,
ymin=82.7566774971172, ymax=95.7302637984316,
ytick style={color=black}
]
\path [draw=blue, fill=blue, opacity=0.1, very thin]
(axis cs:1,84.1780940346412)
--(axis cs:1,83.3463859653588)
--(axis cs:2,88.168295206387)
--(axis cs:3,89.6899914955071)
--(axis cs:4,90.3893580515406)
--(axis cs:5,91.0597924753775)
--(axis cs:6,91.4267411545873)
--(axis cs:7,91.6628142662254)
--(axis cs:8,91.8241609003782)
--(axis cs:9,91.9644454770871)
--(axis cs:10,92.0924721541854)
--(axis cs:11,92.1584019660877)
--(axis cs:12,92.2613349538849)
--(axis cs:13,92.3805725728416)
--(axis cs:14,92.4372551665221)
--(axis cs:15,92.4731079220801)
--(axis cs:16,92.5808221276101)
--(axis cs:17,92.5654972980607)
--(axis cs:18,92.6322550161315)
--(axis cs:19,92.6523837094094)
--(axis cs:20,92.6948032368616)
--(axis cs:20,92.9873567631384)
--(axis cs:20,92.9873567631384)
--(axis cs:19,92.9551362905906)
--(axis cs:18,92.9599849838686)
--(axis cs:17,93.0081827019392)
--(axis cs:16,92.9335778723899)
--(axis cs:15,92.9081720779199)
--(axis cs:14,92.8869848334779)
--(axis cs:13,92.8885474271584)
--(axis cs:12,92.8512250461151)
--(axis cs:11,92.7319980339124)
--(axis cs:10,92.6064878458145)
--(axis cs:9,92.468034522913)
--(axis cs:8,92.3187190996218)
--(axis cs:7,92.0418257337746)
--(axis cs:6,91.7713388454127)
--(axis cs:5,91.4726075246225)
--(axis cs:4,90.9356019484594)
--(axis cs:3,90.2716885044929)
--(axis cs:2,88.281224793613)
--(axis cs:1,84.1780940346412)
--cycle;

\path [draw=red, fill=red, opacity=0.1, very thin]
(axis cs:0,85.4248)
--(axis cs:0,85.4248)
--(axis cs:1,88.7191216145834)
--(axis cs:2,90.2482331266916)
--(axis cs:3,90.7726259894957)
--(axis cs:4,91.1549598808515)
--(axis cs:5,91.5257626910518)
--(axis cs:6,91.7203920528709)
--(axis cs:7,91.9091154317412)
--(axis cs:8,92.005969913527)
--(axis cs:9,92.1120473464203)
--(axis cs:10,92.203501230955)
--(axis cs:11,92.2539246128922)
--(axis cs:12,92.3368874149428)
--(axis cs:13,92.4247302127112)
--(axis cs:14,92.4820129537943)
--(axis cs:15,92.5041122814274)
--(axis cs:16,92.6129963446373)
--(axis cs:17,92.5935286904868)
--(axis cs:18,92.6527233742261)
--(axis cs:19,92.6731827610205)
--(axis cs:20,92.7196380014973)
--(axis cs:20,92.9762019985027)
--(axis cs:20,92.9762019985027)
--(axis cs:19,92.9418572389795)
--(axis cs:18,92.9515166257739)
--(axis cs:17,92.9955113095132)
--(axis cs:16,92.9302036553626)
--(axis cs:15,92.9108477185726)
--(axis cs:14,92.8897470462057)
--(axis cs:13,92.8948697872888)
--(axis cs:12,92.8399925850572)
--(axis cs:11,92.7472753871078)
--(axis cs:10,92.645458769045)
--(axis cs:9,92.5061126535796)
--(axis cs:8,92.402670086473)
--(axis cs:7,92.2192045682588)
--(axis cs:6,92.0552879471291)
--(axis cs:5,91.8603173089482)
--(axis cs:4,91.5270401191485)
--(axis cs:3,91.2720940105043)
--(axis cs:2,90.4146468733084)
--(axis cs:1,89.2395183854167)
--(axis cs:0,85.4248)
--cycle;

\path [draw=green, fill=green, opacity=0.1, very thin]
(axis cs:1,87.1339902576174)
--(axis cs:1,86.1520097423826)
--(axis cs:2,90.7019929214453)
--(axis cs:3,92.2857090704337)
--(axis cs:4,92.8763483866582)
--(axis cs:5,93.478821461305)
--(axis cs:6,93.7920054906926)
--(axis cs:7,94.0826655067383)
--(axis cs:8,94.1955547319614)
--(axis cs:9,94.2853199487134)
--(axis cs:10,94.4075178731339)
--(axis cs:11,94.4761092785948)
--(axis cs:12,94.5557766066365)
--(axis cs:13,94.6267517191125)
--(axis cs:14,94.6531223911986)
--(axis cs:15,94.6633405610935)
--(axis cs:16,94.7738128272028)
--(axis cs:17,94.7367246698099)
--(axis cs:18,94.7897959845778)
--(axis cs:19,94.804959176615)
--(axis cs:20,94.8543269412026)
--(axis cs:20,95.0992730587974)
--(axis cs:20,95.0992730587974)
--(axis cs:19,95.093280823385)
--(axis cs:18,95.1034840154222)
--(axis cs:17,95.14055533019)
--(axis cs:16,95.0735471727972)
--(axis cs:15,95.0580994389065)
--(axis cs:14,95.0596776088014)
--(axis cs:13,95.0499682808875)
--(axis cs:12,95.0171833933635)
--(axis cs:11,94.9117307214052)
--(axis cs:10,94.7998421268661)
--(axis cs:9,94.7285200512866)
--(axis cs:8,94.5788452680386)
--(axis cs:7,94.3884544932617)
--(axis cs:6,94.1644745093074)
--(axis cs:5,93.787578538695)
--(axis cs:4,93.2738116133418)
--(axis cs:3,92.6899709295663)
--(axis cs:2,91.0448870785548)
--(axis cs:1,87.1339902576174)
--cycle;

\path [draw=orange, fill=orange, opacity=0.1, very thin]
(axis cs:0,88.135)
--(axis cs:0,88.135)
--(axis cs:1,91.5099586644598)
--(axis cs:2,92.8534252376527)
--(axis cs:3,93.3270789870073)
--(axis cs:4,93.5993625352275)
--(axis cs:5,93.961638667655)
--(axis cs:6,94.1011844287411)
--(axis cs:7,94.2714405190698)
--(axis cs:8,94.337436838261)
--(axis cs:9,94.403170946109)
--(axis cs:10,94.4886105952307)
--(axis cs:11,94.5355530293236)
--(axis cs:12,94.6050173793674)
--(axis cs:13,94.6751212211024)
--(axis cs:14,94.6908665313625)
--(axis cs:15,94.6871071470993)
--(axis cs:16,94.7962436271194)
--(axis cs:17,94.7695811373299)
--(axis cs:18,94.826417572806)
--(axis cs:19,94.804959176615)
--(axis cs:20,94.8773285164322)
--(axis cs:20,95.0917914835678)
--(axis cs:20,95.0917914835678)
--(axis cs:19,95.093280823385)
--(axis cs:18,95.073342427194)
--(axis cs:17,95.1189788626701)
--(axis cs:16,95.0552763728807)
--(axis cs:15,95.0650528529007)
--(axis cs:14,95.0558534686375)
--(axis cs:13,95.0436787788976)
--(axis cs:12,95.0177826206326)
--(axis cs:11,94.9366069706764)
--(axis cs:10,94.8426694047693)
--(axis cs:9,94.767069053891)
--(axis cs:8,94.674323161739)
--(axis cs:7,94.5023194809302)
--(axis cs:6,94.3744155712589)
--(axis cs:5,94.144761332345)
--(axis cs:4,93.8867974647725)
--(axis cs:3,93.7070810129927)
--(axis cs:2,93.0308147623473)
--(axis cs:1,92.1568413355402)
--(axis cs:0,88.135)
--cycle;

\addplot [ultra thick, blue]
table {%
0 nan
1 83.76224
2 88.22476
3 89.98084
4 90.66248
5 91.2662
6 91.59904
7 91.85232
8 92.07144
9 92.21624
10 92.34948
11 92.4452
12 92.55628
13 92.63456
14 92.66212
15 92.69064
16 92.7572
17 92.78684
18 92.79612
19 92.80376
20 92.84108
};
\addlegendentry{Baseline}
\addplot [ultra thick, red]
table {%
0 85.4248
1 88.97932
2 90.33144
3 91.02236
4 91.341
5 91.69304
6 91.88784
7 92.06416
8 92.20432
9 92.30908
10 92.42448
11 92.5006
12 92.58844
13 92.6598
14 92.68588
15 92.70748
16 92.7716
17 92.79452
18 92.80212
19 92.80752
20 92.84792
};
\addlegendentry{Mean from text} 
\addplot [ultra thick, green]
table {%
0 nan
1 86.643
2 90.87344
3 92.48784
4 93.07508
5 93.6332
6 93.97824
7 94.23556
8 94.3872
9 94.50692
10 94.60368
11 94.69392
12 94.78648
13 94.83836
14 94.8564
15 94.86072
16 94.92368
17 94.93864
18 94.94664
19 94.94912
20 94.9768
};
\addlegendentry{Covariance from text} 
\addplot [ultra thick, orange]
table {%
0 88.135
1 91.8334
2 92.94212
3 93.51708
4 93.74308
5 94.0532
6 94.2378
7 94.38688
8 94.50588
9 94.58512
10 94.66564
11 94.73608
12 94.8114
13 94.8594
14 94.87336
15 94.87608
16 94.92576
17 94.94428
18 94.94988
19 94.94912
20 94.98456
};
\addlegendentry{Mean and covariance from text} 
\end{axis}
\end{scope}
\end{tikzpicture}}}}\hfil 
\subfloat[Multi-class]{\scalebox{0.76}{{
\begin{tikzpicture}

\definecolor{green}{RGB}{0,128,0}
\definecolor{lightgray203}{RGB}{203,203,203}
\definecolor{lightgray204}{RGB}{204,204,204}
\definecolor{orange}{RGB}{255,165,0}
\definecolor{whitesmoke240}{RGB}{240,240,240}

\begin{axis}[
axis background/.style={fill=white},
axis line style={whitesmoke240},
legend cell align={left},
legend style={
  fill opacity=0.8,
  draw opacity=1,
  text opacity=1,
  at={(0.97,0.03)},
  anchor=south east,
  draw=lightgray204,
  fill=white
},
tick align=outside,
tick pos=left,
unbounded coords=jump,
x grid style={lightgray203},
xlabel={number of shots},
xmajorgrids,
xmin=-0.8, xmax=16.8,
xtick style={color=black},
y grid style={lightgray203},
ylabel={Accuracy (\%)},
ymajorgrids,
ymin=25.8988898992538, ymax=80.4112035036087,
ytick style={color=black}
]

\addplot [ultra thick, blue]
table {%
0 nan
1 37.8743797540665
2 52.1549761295319
3 58.7150037288666
4 63.1336987018585
5 66.0803377628326
6 67.5512373447418
7 69.3498969078064
8 70.5901026725769
9 71.3703751564026
10 72.182160615921
11 72.7397441864014
12 73.1344878673553
13 73.7253248691559
14 73.9658176898956
15 74.4685173034668
16 74.6524095535278
};
\addlegendentry{Baseline}
\addplot [ultra thick, red]
table {%
0 28.3767223358154
1 45.1527386903763
2 55.6102216243744
3 60.6912076473236
4 64.4045531749725
5 66.8761134147644
6 68.195503950119
7 69.704270362854
8 70.7999527454376
9 71.6457784175873
10 72.3379969596863
11 72.8130877017975
12 73.0827152729034
13 73.7753689289093
14 73.954564332962
15 74.4476079940796
16 74.6524095535278
};
\addlegendentry{Mean from text}
\addplot [ultra thick, green]
table {%
0 nan
1 40.6697183847427
2 56.5455377101898
3 63.6779427528381
4 68.0416882038116
5 70.7603693008423
6 72.0561742782593
7 73.5299646854401
8 74.6313631534576
9 75.1939117908478
10 75.995671749115
11 76.2982606887817
12 76.7946898937225
13 77.3121118545532
14 77.4996757507324
15 77.7779698371887
16 77.7579188346863
};
\addlegendentry{Covariance from text}
\addplot [ultra thick, orange]
table {%
0 32.6025813817978
1 49.5404928922653
2 61.0295653343201
3 66.1261975765228
4 69.6146249771118
5 71.6841399669647
6 72.6912081241608
7 73.885190486908
8 75.0558137893677
9 75.6368398666382
10 76.1121451854706
11 76.6123175621033
12 76.8548309803009
13 77.3474514484406
14 77.5876581668854
15 77.9090702533722
16 77.9333710670471
};
\addlegendentry{Mean and covariance from text}

\addplot [only marks, purple, mark=diamond*, mark size=3.5pt]
table {
0 29.94
};
\addlegendentry{Zero-Shot CLIP}
\end{axis}

\end{tikzpicture}}}}\hfil 
\vspace{-0mm}
\caption{Results on iNaturalist Dataset for both one-class and multi-class classification tasks using $10^3$ few-shot runs. For all the models we use a Mahalanobis classifier. We compare the baseline without text to three models. 
}
\label{fig:inaturalist}
\end{figure}
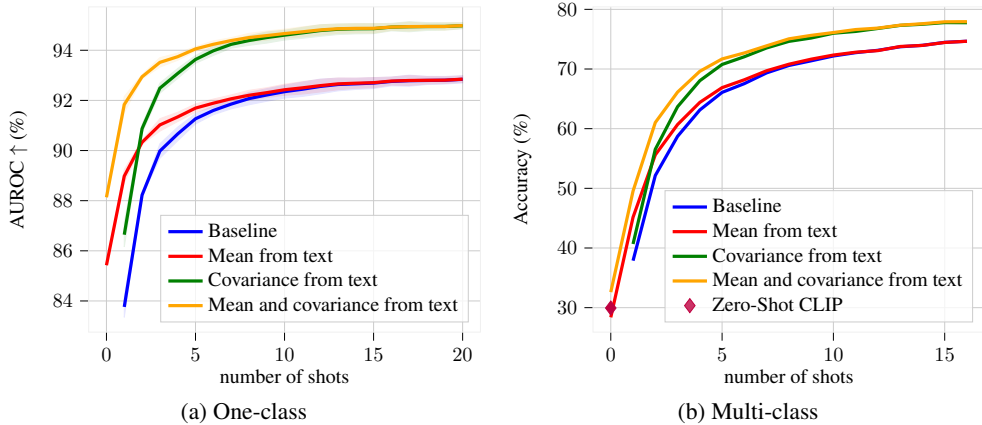
Next, we use the mean and covariance predictions from text for few-shot classification tasks and show that these bring consistent improvements over the existing baselines in both one-class and multi-class scenarios. We start from a baseline which uses only the few available shots with no text supervision i.e. $\alpha=0$ and $\beta=0$. We compare the baseline to three models: M (mean from the shots and text) i.e. $\beta=0$, C (covariance from the text) i.e. $\alpha=0$ and M\&C (mean and covariance from text). For the multi-class setting, we select the best $\alpha$ and $\beta$ coefficient using a held-out validation set with $min(n,4)$ examples, where n is the number of shots, which is a common practice in the field~\cite{coop}.

\subsubsection{One-class classification}
Let us first consider the one-class classification case. In this setting, the goal is to determine if test samples belong to a specific class or not. It contrasts with multi-class. In this context, the covariance matrix becomes pivotal. Using the Mahalanobis distance, the sample likelihood is: 
\begin{equation}
p(x | \mu, \Sigma) = \frac{1}{\sqrt{(2\pi)^d |\Sigma|}} \exp\left(-\frac{1}{2} (x - \mu)^T \Sigma^{-1} (x - \mu)\right)
\end{equation}

We report the AUROC (Area Under the ROC Curve) score for different methods in Figure~\ref{fig:inaturalist} for iNaturalist in-domain and Table~\ref{table:cross_domain_openset_no_deltas} for cross-domain datasets. For each run, we select a random class in the test set with $k$ random shots from the corresponding class and select with a probability of $0.5$ a query either from the same class or from another class.
For iNaturalist, our method shows a consistent improvement over the baseline. We show that using the mean from text brings an improvement in the very low shot regime ($k\leq10-$shots). While including the covariance brings a consistent improvement of $\simeq 2\%$ for all data regimes. Particularly in a very low shot regime, it's interesting to combine both mean and covariance predictions from text which brings over $8\%$ AUROC improvement. Furthermore, our method allows to perform zero-shot with performances equivalent to $2-$shots classification.

In the cross-domain setting, Table~\ref{table:cross_domain_openset_no_deltas} shows that for most of the datasets, our method (M\&C) brings a significant improvement over the baseline. Overall, using the covariance seems to bring more improvement than using the mean. This could be explained by the importance of the covariance for the one-class classification task. Similarly to iNaturalist, using the mean helps more in the very low shot regimes and the using covariance helps for all the different number of shots that we tested.

\begin{center}
\newcommand{\uparrowgreen}{\tikz{\draw[green,->] (0,0) -- ++(0,0.3);}}
\newcommand{\downarrowred}{\tikz{\draw[red,->] (0,0.3) -- ++(0,-0.3);}}
\begin{table}[htbp]
\centering
\scalebox{0.73}{\begin{tabular}{cc|c|ccccccccc|c}
\toprule
& \multirow{2}{*}{Method} & \multirow{2}{*}{Shots} & \multicolumn{10}{c}{Dataset}  \\ 
\cmidrule(r){4-13}
& & & Caltech & EuroSAT & Food & Flowers & SUN397 & DTD & Pets & Cars & UCF101 & Average \\ 

\midrule
\multirow{5}{*}{\rotatebox{90}{AUROC}}&\multirow{5}{*}{\parbox{2cm}{\centering Baseline}} 
& 1 & $89.74$ & $81.34$ & $83.34$ & $90.92$ & $89.18$ & $71.23$ & $81.47$ & $90.31$ & $84.70$ & $84.70$ \\
&& 2 & $93.31$ &  $85.79$& $87.59$ & $94.66$ & $94.37$ & $76.36$ & $86.16$ & $94.28$ & $88.63$ & $89.02$\\
&& 4 & $94.95$ & $88.60$ & $89.54$ & $96.38$ & $96.47$ & $79.53$ &  $88.11$ & $96.04$ & $90.70$ & $91.15$ \\
&& 8 & $95.89$ & $89.85$ & $90.36$ & $97.09$ & $97.51$ & $80.88$ & $88.83$ & $96.78$ & $91.72$ &  $92.10$\\
&& 16 & $96.49$ & $90.39$ &$90.79$  & $97.51$ & $97.83$ & $81.63$ &  $89.21$ & $97.12$ & $92.25$ & $92.57$ \\
\midrule
\multirow{14}{*}{\rotatebox{90}{$\Delta\%$}}&\multirow{5}{*}{\parbox{2cm}{\centering Mean from text (M)}} 
& 1 & +$2.07$ & $0.00$ & +$0.68$ & +$0.48$ & +$2.83$ & +$0.17$ &  +$6.31$ & +$0.38$ & +$1.37$ & $1.59$\\ 
&& 2 & +$0.46$ & $0.00$ & $0.00$ & $0.00$ & +$0.22$ & $0.00$ &  +$2.17$ & $0.00$ & +$0.11$ &  $0.33$\\ 
&& 4 & +$0.10$ & $0.00$ & $0.00$ & $0.00$ & $0.00$ & $0.00$ &  +$0.76$ & $0.00$ & $0.00$ &  $0.10$\\ 
&& 8 & $0.00$ & $0.00$ & $0.00$ & $0.00$ & $0.00$ & $0.00$ &  +$0.37$ & $0.00$ & $0.00$ & $0.04$ \\ 
&& 16 & $0.00$ & $0.00$ & $0.00$ & $0.00$ & $0.00$ & $0.00$ & +$0.18$ & $0.00$ & $0.00$ & $0.02$ \\ 
\cmidrule(r){2-13}
& \multirow{5}{*}{\parbox{2cm}{\centering Covariance from text (C)}} 
& 1 & +$2.09$ & +$1.47$ & +$3.05$ & +$3.59$ & +$0.08$ & $0.00$ & +$4.57$ & +$1.97$ & $0.00$ &  +$1.86$\\ 
&& 2 & +$1.80$ & +$1.45$ & +$3.09$ & +$2.53$ & +$0.21$ & $0.00$ &  +$3.38$ & +$1.66$ & $0.00$ &  +$1.57$\\ 
&& 4 & +$\mathbf{1.49}$ & +$\mathbf{1.27}$ & +$\mathbf{2.92}$ & +$\mathbf{1.82}$ & +$\mathbf{0.21}$ & $\mathbf{0.00}$ &  +$\mathbf{2.76}$ & +$\mathbf{1.23}$ & $\mathbf{0.00}$ & +$\mathbf{1.30}$ \\ 
&& 8 & +$\mathbf{1.15}$ & +$\mathbf{1.22}$ & +$\mathbf{2.73}$ & +$\mathbf{1.45}$ & +$\mathbf{0.22}$ & $\mathbf{0.00}$ &  +$\mathbf{2.53}$ & +$\mathbf{1.07}$ & $\mathbf{0.00}$ & +$\mathbf{1.15}$ \\ 
&& 16 & +$\mathbf{0.96}$ & +$\mathbf{1.20}$ & +$\mathbf{2.64}$ & +$\mathbf{1.16}$ & +$\mathbf{0.25}$ & $\mathbf{0.00}$ & +$\mathbf{2.40}$ & +$\mathbf{1.00}$ & $\mathbf{0.00}$ & +$\mathbf{1.07}$\\  

\cmidrule(r){2-13}
&\multirow{5}{*}{\parbox{2cm}{\centering Mean and covariance from text (M\&C)}} 
& 1 & +$\mathbf{3.82}$ & +$\mathbf{1.47}$ & +$\mathbf{3.61}$ & +$\mathbf{3.99}$ & +$\mathbf{2.85}$ & +$\mathbf{0.17}$ &  +$\mathbf{8.43}$ & +$\mathbf{2.33}$ & +$\mathbf{1.37}$ & +$\mathbf{3.11}$ \\ 
&& 2 & +$\mathbf{2.07}$ & +$\mathbf{1.45}$ & +$\mathbf{3.09}$ & +$\mathbf{2.54}$ & +$\mathbf{0.34}$ & $0.00$ & +$\mathbf{4.18}$ & +$\mathbf{1.66}$ & +$\mathbf{0.11}$ & +$\mathbf{1.71}$ \\ 
&& 4 & +$\mathbf{1.52}$ & +$\mathbf{1.27}$ & +$\mathbf{2.92}$ & +$\mathbf{1.82}$ & +$\mathbf{0.21}$ & $0.00$ &  +$\mathbf{2.79}$ & +$\mathbf{1.23}$ & +$0.01$ & +$\mathbf{1.31}$ \\ 
&& 8 & +$\mathbf{1.15}$ & +$\mathbf{1.22}$ & +$\mathbf{2.73}$ & +$\mathbf{1.45}$ & +$\mathbf{0.22}$ & $0.00$ &  +$\mathbf{2.53}$ & +$\mathbf{1.07}$ & $0.00$ & +$\mathbf{1.15}$ \\ 
&& 16 & +$\mathbf{0.96}$ & +$\mathbf{1.20}$ & +$\mathbf{2.64}$ & +$\mathbf{1.16}$ & +$\mathbf{0.25}$ & $0.00$ & +$\mathbf{2.40}$ & +$\mathbf{1.00}$ & $0.00$ & +$\mathbf{1.07}$ \\ 

\bottomrule
\end{tabular}
}
\vspace{2mm}
\caption{Results per dataset on one-class tasks. We compute the AUROC score over 100 tasks and average the performance over 20 runs. We report the score difference with respect to the Baseline for different number of shots. We highlight the best performance gains compared to the baseline. The confidence intervals range from $0.07\%$ to $0.67\%$ which we omitted for better readability.}
\label{table:cross_domain_openset_no_deltas}
\end{table}
\end{center}
\vspace{-1.2cm}
\subsubsection{Multi-class classification}
In the multi-class scenario, our approach utilizes the same Mahanalobis classifier, a generalization of the Nearest Class Mean which has been widely adopted in the few-shot literature~\cite{hu2022pushing,protonet,easy}: 
\begin{equation}
p(y=c/x)=\frac{\exp\left(\sqrt{\left(\mathbf{\hat{\mu}}_c-\mathbf{x}\right)^T\mathbf{\hat{\Sigma}}^{-1}_c\left(\mathbf{\hat{\mu}}_c-\mathbf{x}\right)}\right)}{\sum_{c'}^{} \exp\left(\sqrt{\left(\mathbf{\hat{\mu}}_{c'}-\mathbf{x}\right)^T\mathbf{\hat{\Sigma}}^{-1}_{c'}\left(\mathbf{\hat{\mu}}_{c'}-\mathbf{x}\right)}\right)} 
\end{equation}

For this setting, we report the performance over $10^3$ runs. Similarly to the one-class setting, our baseline benefits from the additional gain from the mean and the covariance. On iNaturalist, the gain in multi-class is smaller, which could be explained by the difficulty of the task and the fact that one-class depends strongly on a good estimate of the boundaries of the distribution which is reflected in the covariance matrix. Furthermore, we observe on Table~\ref{table:cross_domain_multiclass_no_deltas} that the benefits of our methods are particularly important on datasets where the baseline is low compared to the zero-shot performance (Pets and SUN397). In the case where the zero-shot is lower than the baseline, our method still brings improvements, for example we can observe on EuroSAT $2\%$ increase for all shot settings. 
\begin{center}
\newcommand{\uparrowgreen}{\tikz{\draw[green,->] (0,0) -- ++(0,0.3);}}
\newcommand{\downarrowred}{\tikz{\draw[red,->] (0,0.3) -- ++(0,-0.3);}}
\begin{table}[htbp]
\label{table:MSE}
\centering
\scalebox{0.685}{
\begin{tabular}{cc|c|ccccccccc|c}
\toprule
& \multirow{2}{*}{Method} & \multirow{2}{*}{Shots} & \multicolumn{10}{c}{Dataset}  \\ 
\cmidrule(r){4-13}
& & & Caltech & EuroSAT & Food & Flowers & SUN397 & DTD & Pets & Cars & UCF101 & Average \\ 
\cmidrule(r){1-13}
\multicolumn{2}{c}{\parbox{3cm}{\centering Zero-Shot CLIP}} & 0 & 82.39 & 36.01 & 75.72 & 62.28 & 56.60 & 40.01 &  79.59 & 51.91 & 59.42 & 60.43\\ 
\midrule

\multirow{5}{*}{\rotatebox{90}{Accuracy}} & \multirow{5}{*}{\parbox{2cm}{\centering Baseline}} 
& 1 & $62.07$ & $47.80$ & $27.55$ & $52.25$ & $28.15$ & $26.11$ & $26.69$ & $21.62$ & $36.89$ &$36.57$\\
& & 2 & $74.95$ & $58.30$ &$40.33$ &$69.10$ & $41.42$ & $37.64$  & $38.28$ & $33.18$ & $50.08$ &$49.25$\\
& & 4 & $81.31$ & $66.07$ & $51.67$ & $80.49$ & $51.89$ & $47.49$  & $49.65$ & $44.02$ & $58.80$ &$59.04$\\
& & 8 & $84.61$ & $70.67$ & $59.72$ & $86.52$ & $58.59$ & $54.17$  & $58.63$ & $52.53$ & $64.23$ &$65.52$\\
& & 16 & $86.34$ & $73.49$ & $64.68$ & $89.59$ & $62.46$ & $58.21$ & $65.23$ & $65.18$ & $67.48$ &$70.68$\\

\cmidrule(r){1-13}
\multirow{14}{*}{\rotatebox{90}{$\Delta\%$}}  & \multirow{5}{*}{\parbox{2cm}{\centering Mean  from text (M)}} 
& 1 & +$16.46$ & +$0.82$ & +$13.80$ & +$6.58$ & +$17.22$ & +$6.94$  & +$26.56$ & +$5.26$ & +$9.98$ & +$11.51$ \\
&& 2 & +$6.30$ & +$0.33$ & +$8.09$ & +$2.58$ & +$9.00$ & +$3.62$  & +$19.74$ & +$2.04$ & +$4.06$ & +$06.20$ \\
&& 4 & +$2.32$ & -$0.07$ & +$3.85$ & +$0.75$ & +$3.66$ & +$1.49$ & +$13.40$ & +$0.58$ & +$1.53$ & +$03.06$ \\
&& 8 & +$0.76$ & -$0.19$ & +$1.57$ & +$0.20$ & +$1.30$ & +$0.55$ & +$8.91$ & +$0.13$ & +$0.51$ & +$01.53$ \\
&& 16 & +$0.21$ & -$0.33$ & +$0.67$ & +$0.04$ & +$0.43$ & +$0.28$ & +$5.61$ & +$5.76$ & +$0.15$ & +$0.91$ \\
\cmidrule(r){2-13}
&\multirow{5}{*}{\parbox{2cm}{\centering Covariance from text (C)}} 
& 1 & +$1.56$ & +$0.84$ & +$4.06$ & +$6.34$ & +$0.74$ & +$1.79$  & +$6.64$ & +$4.93$ & +$1.48$ & +$03.15$ \\
&& 2 & +$2.92$ & +$1.52$ & +$6.72$ & +$6.84$ & +$2.16$ & +$3.75$  & +$11.10$ & +$7.58$ & +$3.50$ & +$05.12$ \\
&& 4 & +$2.85$ & +$1.97$ & +$7.41$ & +$5.35$ & +$3.21$ & +$5.27$  & +$13.29$ & +$8.72$ & +$4.55$ & +$05.85$ \\
&& 8 & +$2.44$ & +$2.03$ & +$6.71$ & +$3.88$ & +$3.59$ & +$5.44$  & +$12.60$ & +$8.77$ & +$4.98$ & +$05.60$ \\
&& 16 & +$2.29$ & +$1.99$ & +$5.80$ & +$2.95$ & +$3.59$ & +$5.33$ & +$11.31$ & +$11.22$  & +$5.11$ & +$04.82$ \\ 
\cmidrule(r){2-13}
&\multirow{5}{*}{\parbox{2cm}{\centering Mean and covariance from text (M\&C)}} 
& 1 & +$\mathbf{21.03}$ & +$\mathbf{2.20}$ & +$\mathbf{25.57}$ & +$\mathbf{14.90}$ & +$\mathbf{22.14}$ & +$\mathbf{14.05}$  & +$\mathbf{37.86}$ & +$\mathbf{16.14}$ & +$\mathbf{14.94}$ & +$\mathbf{18.76}$ \\
&& 2 & +$\mathbf{10.27}$ & +$\mathbf{2.04}$ & +$\mathbf{18.42}$ & +$\mathbf{9.82}$ & +$\mathbf{13.74}$ & +$\mathbf{10.45}$  & +$\mathbf{31.86}$ & +$\mathbf{11.22}$ & +$\mathbf{9.48}$ & +$\mathbf{13.03}$ \\
&& 4 & +$\mathbf{5.76}$ & +$\mathbf{2.21}$ & +$\mathbf{12.49}$ & +$\mathbf{6.21}$ & +$\mathbf{8.23}$ & +$\mathbf{8.15}$  & +$\mathbf{25.80}$ & +$\mathbf{9.67}$ & +$\mathbf{7.27}$ & +$\mathbf{09.53}$ \\
&& 8 & +$\mathbf{3.71}$ & +$\mathbf{2.09}$ & +$\mathbf{8.78}$ & +$\mathbf{4.21}$ & +$\mathbf{5.56}$ & +$\mathbf{6.71}$ & +$\mathbf{20.42}$ & +$\mathbf{9.06}$ & +$\mathbf{6.29}$ & +$\mathbf{07.43}$ \\
&& 16 & +$\mathbf{2.81}$ & +$\mathbf{2.02}$ & +$\mathbf{6.74}$ & +$\mathbf{3.13}$ & +$\mathbf{4.46}$ & +$\mathbf{6.10}$ & +$\mathbf{16.30}$ & +$\mathbf{15.99}$ & +$\mathbf{5.91}$ & +$\mathbf{5.94}$ \\ 
\bottomrule
\end{tabular}
}
\vspace{2mm}
\caption{Results per dataset. We report the performance across $10^2$ tasks of Zero-shot CLIP and the baseline, for other methods we report the average difference to the baseline. We observe that combining and mean and covariance (M\&C) consistently improves the baseline. The confidence intervals range from $0.09\%$ to $0.35\%$ which we omitted for better readability.}
\label{table:cross_domain_multiclass_no_deltas}
\end{table}

\end{center}

\vspace{-1.4cm}
\section{Discussion}

Let us now summarize and discuss our findings through Q\&As.

\textbf{Q1}: Can we predict the mean and the covariance from text? \textbf{A}: Yes, our experiment shows that we are able to predict the mean and covariance. Our interpretation is that the multi-modal training of CLIP helps aligning the two modalities, therefore training a small MLP network is sufficient to learn the mapping to the mean and covariance.

\textbf{Q2}: Does Covariance bring improvement in few-shot classification? \textbf{A}: Yes, we show that for both one-class and multi-class settings, using the covariance matrix to normalize the data brings an improvement. Especially when dealing with fine-grained concepts such as species from iNaturalist.

\textbf{Q3}: Which form of covariance should we use? \textbf{A}: We use a diagonal form for the covariance matrix given the high dimension of the features (1024) and the limited number of samples in the base datasets. Even when using ImageNet, computing a full covariance matrix is undetermined. Reducing the dimensionality of the features with methods such as PCA led to poor results.

\textbf{Q4}: How important is the choice of $\alpha$ and $\beta$? \textbf{A}: Our results show that the importance of $\alpha$ and $\beta$ varies depending on two things. The first one is the class setting:  for one-class tasks, we observed that the best hyperparameters are $\alpha=0.1$ and $\beta=1$ which confirms the importance of the covariance matrix in this task. For the multi-class setting, we obtain different best hyperparamters depending on the task. We use the validation set on each task to select these hyperparameters. The second factor is the domain shift, we also observed that the importance of $\alpha$ and $\beta$ vary depending on the domain gap between the base and test datasets.


\section{Conclusion}
In conclusion, our work takes a step toward more fully integrating text as a modality in few-shot learning, focusing on its ability to predict mean and covariance statistics for visual feature distributions. The preliminary results demonstrate improvements in robustness and generalizability across different datasets, without claiming a definitive solution to the challenges in the field. For future work, exploring higher-order moments and investigating the applicability of our approach in tasks other than classification could offer further insights into the synergistic use of text and visual data. 

\section*{References}
\medskip
  {\small
\bibliographystyle{unsrtnat}
\renewcommand{\bibsection}{}
\bibliography{egbib}
}
\newpage

\appendix
\section{Supplementary Material}

\subsection{Data collection}
We use instruct-GPT~\cite{gpt3} to generate a visual description of each class. We use different prompts depending on the dataset:

\begin{table}[h]
    \centering
    \begin{tabular}{ll}
        \toprule
        Dataset & Prompt Used \\
        \midrule
        iNaturalist & How can you identify a \{class\} \\
        Pets & \parbox{10cm}{How can you identify a \{class\}?}
         \\
        SUN397 & How can you identify a \{class\}? \\
        Eurosat & How can you identify a satellite image of a \{class\}? \\
        DTD & How can you identify a \{class\} texture? \\
        Flowers & How can you identify a \{class\} flower? \\
        Caltech101 & How can you identify a \{class\}? \\
        Food101 & How can you identify a \{class\}? \\
        Stanford Cars & How can you identify a \{class\} car? \\
        UCF101 & How can you identify a \{class\}? \\
        Imagenet & How can you identify a \{class\}? \\
        \bottomrule
    \end{tabular}
    \vspace{2mm}
    \caption{Datasets and Their Corresponding Prompts}
    \label{tab:datasets}
\end{table}

\subsection{Normalization of the input}
In this section, we detail how the normalization is performed. On the base classes, we compute the mean of the input texts and targets i.e. class means and covariances respectively. After that, we apply a Z-score normalization as follows: 

\begin{equation}
    \forall x\in\mathbb{R}^{Nd}: Z_{i,j} = \frac{x_{ij}-\hat{\mu}_j}{\hat{\sigma}_j}
\end{equation}
where $i$ is the $i$-th datapoint and $j$ is the $j$-th feature.
\end{document}